%% file: main.tex
\documentclass{article}

\usepackage{microtype}
\usepackage{graphicx}
\usepackage{subfigure}
\usepackage{booktabs} 

\usepackage{xcolor}
\usepackage{float}
\usepackage{amssymb}
\usepackage{amsthm}

\usepackage{tabularx}
\usepackage{mathtools}
\usepackage{bbm}
\usepackage{comment}
\usepackage{booktabs}
\usepackage{multirow}
\usepackage[percent]{overpic}

\usepackage{amsthm,amsmath,amssymb, amsfonts}
\newtheorem{theorem}{Theorem}[section]
\newtheorem{lemma}[theorem]{Lemma}

\newtheorem{definition}[theorem]{Definition}

\usepackage{hyperref}

\usepackage[preprint]{corl_2021} 

\title{Unlocking Pixels for Reinforcement Learning via Implicit Attention}

\author{
  Krzysztof Marcin Choromanski\thanks{Equal contribution}\\
  Google\\
   \And
   Deepali Jain$^{*}$ \\
   Google \\
   \And
   Wenhao Yu$^{*}$\\
   Google \\
   \And
   Jack Parker-Holder \\
   University of Oxford \\
   \And
   Tingnan Zhang \\
   Google \\
   \And
   Xingyou Song \\
   Google \\
   \And
   Valerii Likhosherstov \\
   University of Cambridge \\
   \And
   Anirban Santara \\
   Google \\
   \And
   Aldo Pacchiano \\
   UC Berkeley\\
   \And
   Yunhao Tang\\
   Columbia University\\
   \And
   Jie Tan \\
   Google\\
   \And
   Adrian Weller\\
   University of Cambridge\\
}

\begin{document}
\maketitle

\begin{abstract}
There has recently been significant interest in training reinforcement learning (RL)  agents in vision-based environments. This poses many challenges, such as high dimensionality and the potential for observational overfitting through spurious correlations. A promising approach to solve both of these problems is an \emph{attention bottleneck}, which provides a simple and effective framework for learning high performing policies, even in the presence of distractions. However, due to poor scalability of attention architectures, these methods cannot be applied beyond low resolution visual inputs, using large patches (thus small attention matrices). In this paper we make use of new efficient attention algorithms, recently shown to be highly effective for Transformers, and demonstrate that these techniques can be successfully adopted for the RL setting. 
This allows our attention-based controllers to scale to larger visual inputs, and facilitate the use of smaller patches, even individual pixels, improving generalization. We show this on a range of tasks from the Distracting Control Suite to vision-based quadruped robots locomotion. We provide rigorous theoretical analysis of the proposed algorithm.
\end{abstract}



\input{intro}

\input{related_work}
\input{algorithm}

\input{theory}
\input{experiments}
\input{conclusion}

\bibliography{refs.bib}

\onecolumn
\appendix

\input{appendix}

\end{document}

%% file: intro.tex
\section{Introduction}

Reinforcement learning (RL \cite{suttonbarto}) considers the problem of an agent learning 
from interactions to maximize reward. Since the introduction of deep neural networks, the field of deep RL has achieved tremendous results, from games \cite{alphago}, to robotics \cite{rubics_cube} and even real world problems \cite{loon}. 

As RL continues to be tested in more challenging settings, there has been increased interest in learning from vision-based observations \cite{planet, slac, dreamer, rad, curl, drq}. This presents several challenges, as not only are image-based observations significantly larger, but they also contain greater possibility of containing confounding variables, which can lead to overfitting \cite{Song2020Observational}. 

A promising approach for tackling these challenges is through the use of \emph{bottlenecks}, which force agents to learn from a low dimensional feature representation. This has been shown to be useful for both improving scalability \cite{planet, dreamer} and generalization \cite{ibac_sni}. In this paper, we focus on \emph{attention bottlenecks}, using an attention mechanism to select the most important regions of the state space or for learning compact latent image encodings. Recent work showed a specific form of hard attention combined effectively with neuroevolution to create agents with significantly fewer parameters and strong generalization capabilities \cite{yujintang}, while also producing interpretable policies.

However, the current form of selective attention proposed is severely limited. It makes use of the most prominent softmax attention mechanism, popularized by \cite{vaswani}, which suffers from quadratic complexity in the size of the attention matrix (i.e. the number of patches). This means that models become significantly slower as vision-based observations become higher resolution, and the effectiveness of the bottleneck is reduced by relying on larger patches.

\begin{figure}[h]
\centering
    \includegraphics[keepaspectratio, width=0.49\textwidth]{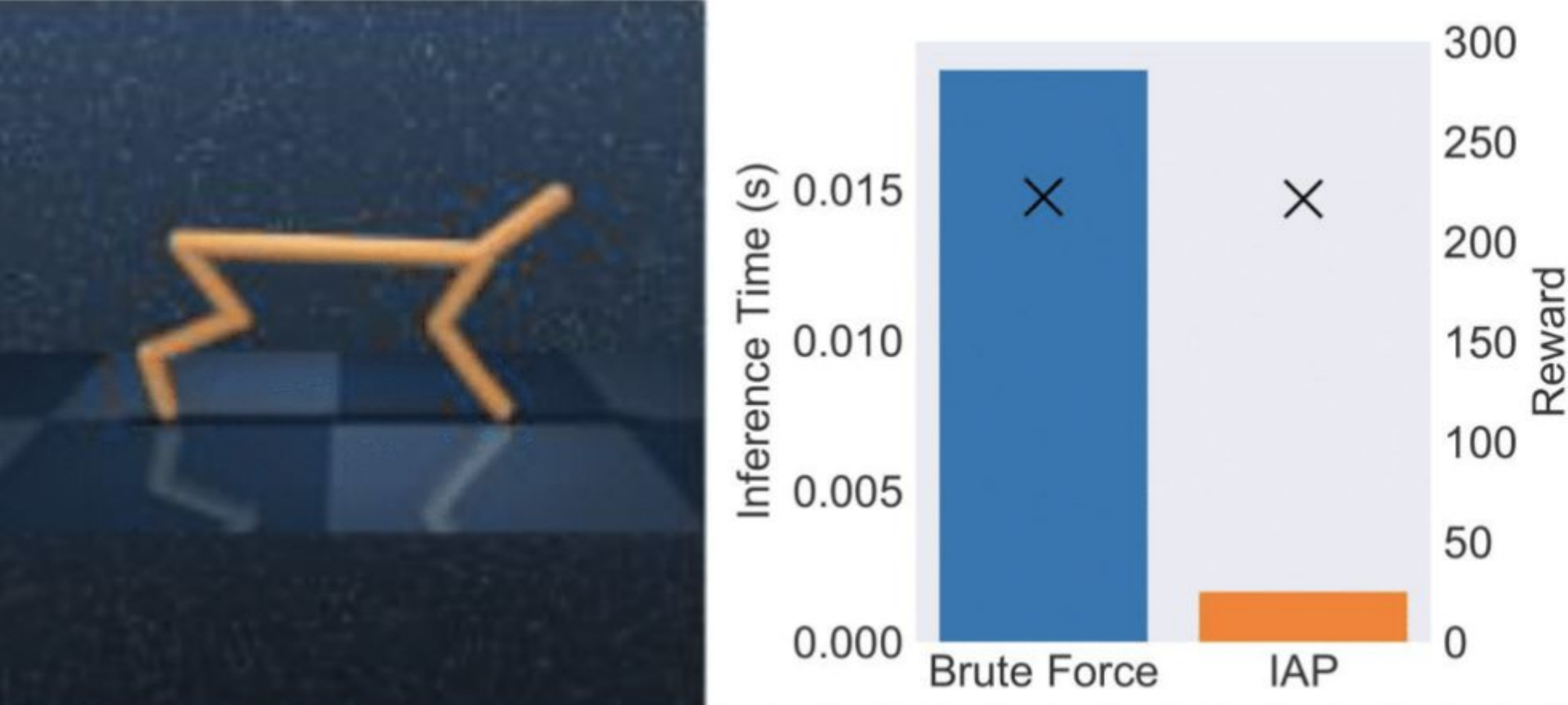}
\vspace{-2mm}
\caption{\small{First plot: An observation from the $\mathrm{Cheetah}$-$\mathrm{Run}$ task, when downsized to a (100 x 100) RGB image. Second plot: comparison of inference time (bars) vs. rewards (crosses) for the Baseline Attention Agent from \cite{yujintang}, and our IAP-rank mechanism. Rewards are the means from five seeds. Policies were trained for 100 iterations. Inference times are the means of 100 forward passes. }}
\label{fig:intro} 
\vspace{-4mm}
\end{figure}


In this paper, we demonstrate how new, scalable attention mechanisms \cite{performers} designed for Transformers can be effectively adapted to the vision-based RL setting. We call the resulting algorithm \textit{Implicit Attention for Pixels} (or IAP). Notably, using IAP we are able to train agents with self-attention for images with 8x more pixels than \cite{yujintang}. We are also able to dramatically reduce the patch size, to even just a single pixel. We show a simple example of the effectiveness of our approach in Figure \ref{fig:intro}. Here we train an agent on the $\mathrm{Cheetah}$:$\mathrm{Run}$ task from the DM Control Suite \cite{dm_control}. All agents are trained in the same way, with the only difference being the attention mechanism used. This leads to dramatically different inference times (in IAP's favor), with IAP policies matching or even outperforming their regular attention counterparts accuracy-wise.

In addition, we show that those IAP variants that leverage random features are effective for RL tasks with as few as 15 random projections, which is in striking contrast to the supervised setting, where usually 200-300 projections are required \citep{performers}. This 13x+ reduction has a profound effect on the speed of the method. IAP comes in two main classes: (a) using attention to \textit{rank} patches based on their importance (\textbf{IAP-rank}); and (b) applying it the same way as in Transformers pipelines (\textbf{IAP-trans}).  
Different IAP variants within each class vary by the attention kernel used and the particular (potentially random) feature mechanism applied for that kernel. They have different strengths (see detailed discussion in Sec. \ref{sec:patctopatch}), with softmax features being more expressive but not always needed.

To summarize, our key contributions are as follows. \textbf{Practical}: To the best of our knowledge, we are the first to use efficient attention mechanisms for RL from pixels. This has two clear benefits: \textbf{1)} we can scale to larger images than previous works; \textbf{2)} we can use more fine-grained patches which produce more effective attention bottlenecks. Both goals can be achieved with an embarrassingly small number of trainable parameters for vision policies (from \textbf{4x} to \textbf{8x} compression over standard CNN-based policies with no loss of quality of the learned controller or even improved accuracy due to easier optimization landscape for the resulting more compact policies, see: Sec. \ref{sec:quadruped_locomotion}). Furthermore, IAP-rank provides \textbf{interpretability} since those regions of the image that an agent pays attention to can be easily visualized and IAP-trans achieves image \textbf{compression}. We test IAP on a rich set of tasks including: Distracting Control Suite \cite{distracting}, navigation, obstacle avoidance and a quadruped robot walking on uneven terrains. \textbf{Theoretical}: We provide rigorous theoretical analysis of IAP. We show that linearization of the attention combined with ranking can leverage a rich class of algorithms on dot-product maximization and nearest neighbor search. We also quantify the quality of those IAP variants that apply approximate softmax kernel estimation techniques.

%% file: related_work.tex
\vspace{-2.5mm}
\section{Related Work}
\vspace{-2.5mm}
Several approaches to vision in RL have been proposed over the years, tackling three key challenges: high-dimensional input space, partial observability of the actual state from images, and observational overfitting to spurious features \cite{Song2020Observational}. Dimensionality reduction can be obtained with hand-crafted features or with learned representations, typically via ResNet/CNN-based modules \citep{resnets}. Other approaches equip an agent with segmentation techniques and depth maps \citep{segmentation}. Those methods require training a substantial number of parameters, just to process vision, usually a part of the richer heterogeneous agent's input, that might involve in addition lidar data, tactile sensors and more as in robotics applications. Partial observability was addressed by a line of work focusing on designing new compact and expressive neural networks for vision-based controllers such as \citep{kulhanek}.

Common ways to reduce observational overfitting are: data augmentation \cite{drq, rad, curl}, causal approaches \cite{zhang2021invariant} and bottlenecks \cite{ibac_sni}. \emph{Information bottlenecks} have been particularly popular in vision-based RL \cite{planet, dreamer, slac}, backed by theory for improved generalization \cite{SHAMIR20102696, 7133169}. 

The idea of selecting individual ``glimpses'' with attention was first proposed by \citet{rnn_visual_attn}, who applied REINFORCE \cite{reinforce} to learn which patches to use, achieving strong generalization results. Others have presented approaches to differentiate through hard attention \cite{bengio2013estimating}. This work is inspired by \citet{yujintang} who proposed to use neuroevolution methods to optimize a hard attention module, circumventing the requirement to backpropagate through it. 

Our paper is also related to the recent line of work on fast attention mechanisms. Since Transformers were shown to produce state-of-the-art results for language modelling tasks \cite{vaswani}, there has been a series of efforts to reduce the $O(L^2)$ time and space with respect to sequence length \cite{Kitaev2020Reformer, peng2021random, wang2020linformer}. This work leverages techniques from Performer architectures \cite{performers}, which were recently shown to be among the best performing efficient mechanisms \cite{tay2021long}, and are well aligned with recent efforts on linearizing attention, exemplified by Performers and LambdaNetworks \cite{irwan}. 


Solving robotics tasks from vision input is an important and well-researched topic~\cite{kalashnikov2018qt, yahya2017collective, levine2016end, Pan2019ZeroshotIL}. Our robotic experiments focus on learning locomotion and navigation skills from vision for legged robots. In prior work, CNNs have been used to process vision \cite{Pan2019ZeroshotIL, Li2019HRL4INHR, blanc2005indoor}. Here, we use  attention to process image observations and compare our results with CNNs for realistic robotics tasks.


%% file: algorithm.tex
\section{Implicit Attention for Vision in Reinforcement Learning}
\label{sec:compact}

\subsection{RL with Attention Bottlenecks}
In this paper we train policies $\pi:\mathcal{S} \rightarrow \mathcal{A}$, where state is a RGB(D) representation of the visual input concatenated with data from other sensors (see: Section \ref{sec:exp}).
The vision part of the state is processed by the attention module that constructs its representation for the subsequent layers of the policy.

\subsection{Image Transformation via IAP Attention - Preliminaries}
Consider an image represented as a collection of $L=a \cdot b$ (potentially intersecting) RGB(D)-patches indexed by $i \in \{0,1,...,a-1\},j \in \{0,1,...,b-1\}$ for some $a, b \in \mathbb{N}_{+}$. Denote by $\mathbf{X} \in \mathbb{R}^{L \times c}$ a matrix with vectorized patches as rows (i.e. vectors of RGB(D)-values of all pixels in the patch) and by $\mathbf{X}^{\prime} \in \mathbb{R}^{L \times c}$ its enriched version with patch position encoding (see: \citep{vaswani}) added to the vectorized patch.
Let $\mathbf{V} \in \mathbb{R}^{L \times d_{V}}$ be a matrix of (potentially learned) value vectors corresponding to patches as in the regular attention mechanism \citep{transformer}. 

For $l \leq L$ and $k \in \mathbb{N}_{+}$, we define the following patch-to-patch attention map $\mathbb{R}^{L \times d_{V}} \rightarrow \mathbb{R}^{l \times d_{V}}$:
\begin{equation}
\label{eq:attention}
\mathrm{Att}(\mathbf{V}) = \Xi(\mathbf{P}_{k, L}\mathbf{A}_{\mathrm{K}},\mathbf{V}),    
\end{equation}
where: $\Xi:\mathbb{R}^{k \times L} \times \mathbb{R}^{L \times d_{V}} \rightarrow \mathbb{R}^{l \times d_{V}}$,
$\mathbf{P}_{k,L} \in \mathbb{R}^{k \times L}$ is a (potentially learnable) projection and:
\begin{itemize}
\item $\mathrm{K}:\mathbb{R}^{d} \times \mathbb{R}^{d} \rightarrow \mathbb{R}_{+}$ is a kernel admitting the form: $\mathrm{K}(\mathbf{u},\mathbf{v}) = \mathbb{E}[\phi(\mathbf{u})^{\top}\phi(\mathbf{v})]$ for some (potentially randomized) finite kernel feature map $\phi:\mathbb{R}^{d} \rightarrow \mathbb{R}^{m}$,
\item $\mathbf{A}_{\mathrm{K}} \in \mathbb{R}^{L \times L}$ is the \textit{attention matrix} defined as: $\mathbf{A}_{\mathrm{K}}(i,j) = \mathrm{K}(\mathbf{q}_{i},\mathbf{k}_{j})$ where $\mathbf{q}_{i},\mathbf{k}_{i}$ are the $i^{th}$ rows of matrices $\mathbf{Q}=\mathbf{X^{\prime}}\mathbf{W}_{\mathrm{Q}}$, $\mathbf{K}=\mathbf{X^{\prime}}\mathbf{W}_{\mathrm{K}}$ (queries \& keys), and $\mathbf{W}_{\mathrm{Q}},\mathbf{W}_{\mathrm{K}} \in \mathbb{R}^{c \times d_{QK}}$ are trainable matrices for some $d_{QK} \in \mathbb{N}_{+}$.
\end{itemize}

\begin{figure*}[h]
  \label{fig:benchmark1}
  \centering
    \vspace{-3mm}
	\includegraphics[keepaspectratio, width=0.99\textwidth]{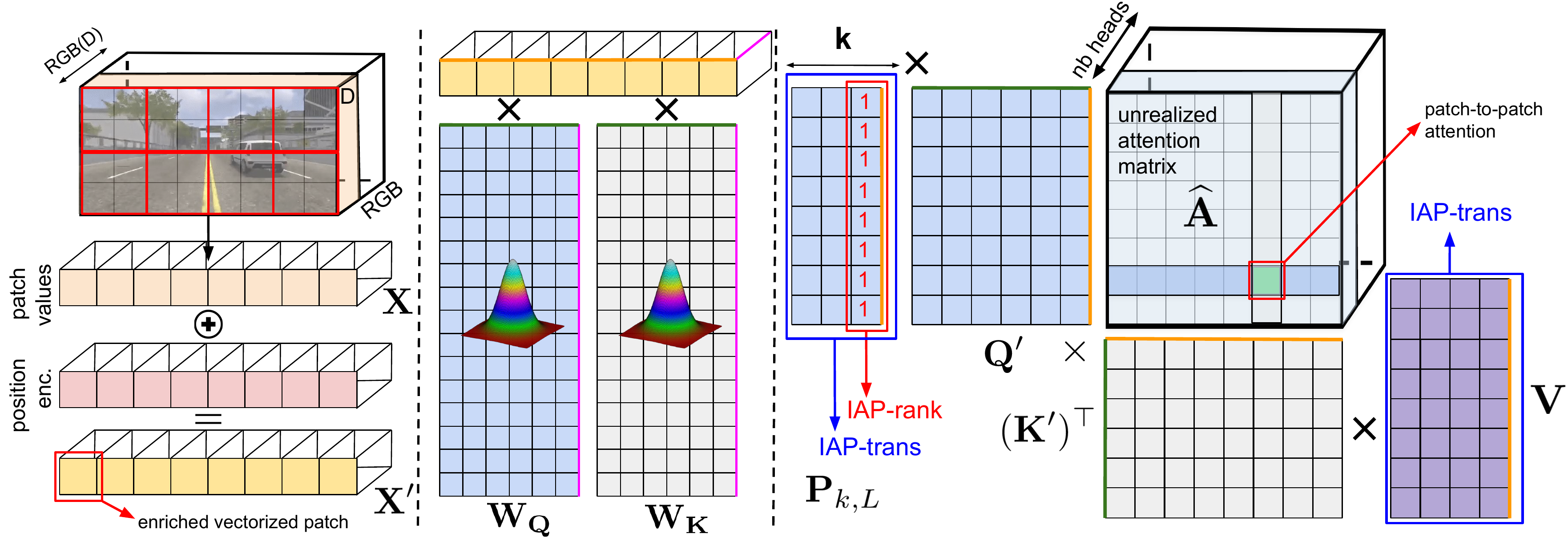}
  \vspace{-4mm}	
  \caption{\small{Visualization of the Implicit Attention for Pixels (IAP). An input RGB(D) is represented as a union of (not necessarily disjoint) patches (in principle even individual pixels). Each patch is projected via learned matrices $\mathbf{W}_{\mathrm{Q}}$/$\mathbf{W}_{\mathrm{K}}$. This is followed by a set of (potentially randomized) projections, which in turn is followed by nonlinear mapping $f$ defining attention type. In the inference, this process can be further optimized by computing the product of $\mathbf{W}_{\mathrm{Q}/\mathrm{K}}$ with the (random) projection matrix in advance. Tensors $\mathbf{Q}^{\prime}$ and $\mathbf{K}^{'}$, obtained via (random) projections followed by $f$, define an attention matrix which is never explicitly materialized. Instead, $\mathbf{Q}^{\prime}$ is left- multiplied by $\mathbf{P}_{k,L}$ (potentially all-one vector if patch score is defined in IAP-rank as average attention value over different queries), the result is right-multiplied by $(\mathbf{K}^{\prime})^{\top}$ and obtained matrix (for IAP-trans) by $\mathbf{V}$ (see: Sec. \ref{sec:patctopatch}).
  The algorithm can in principle use a multi-head mechanism, although we do not apply it in our experiments. Same-color lines indicate axis with the same number of dimensions.}}
  \vspace{-4.5mm}
\label{fig:iap-scheme}  
\end{figure*}
\vspace{-4mm}
\paragraph{IAP-rank:} This is the IAP class with attention defined as in Eq. \ref{eq:attention}, where $k=1$ and vector $\mathbf{r}=\frac{1}{L}\mathbf{P}_{1,L}\mathbf{A}_{\mathrm{K}}$ is interpreted as a vector of scores of different patches. In this setting, function $\Xi$ outputs a masked version of $\mathbf{V}$ with unmasked rows corresponding to patches selected based on the ranking $\mathbf{r}$.
Alternatively (\textit{no masking} version) a sub-matrix consisting of those rows is being output (providing image \textbf{compression}).
The selection can be as simple as taking the top $l$ patches or one can use (for smoothening) softmax sampling based on $\mathbf{r}$ with dot-product kernel $\mathrm{K}$. 
Vector $\mathbf{P}_{1,L}$ defining how scores are computed can be as simple as $\mathbf{P}_{1,L}=[1,...,1]$ (as in \citep{yujintang}) or it can be learnable. The mechanism effectively chooses most important patches of the image
and thus provides as a byproduct \textbf{interpretability} that many other vision-processing systems (e.g. those based on CNN layers) miss.
\vspace{-2mm}
\paragraph{IAP-trans:} For this class, $\mathbf{P}_{k,L}$ for $k=l < L$ is a learnable projection matrix responsible for image \textbf{compression}, and $\Xi$ is defined as:
$\Xi(\mathbf{P}_{k,L}\mathbf{A}_{\mathrm{K}},\mathbf{V})=\mathbf{P}_{k,L}\mathbf{A}_{\mathrm{K}}\mathbf{V}$. The final output of the attention module, as in regular Transformer's attention \citep{vaswani}, is obtained by adding residual connections, and potentially also layer normalization. 

For a summary of both IAP classes, see: Fig. \ref{fig:iap-scheme}.
The output of the attention block is concatenated with signal from other sensors. The former can change in potentially lower frequency than the latter ones, for instance for computational gains or when vision input frequency is different than other. If that is the case, $\mathbf{A}_{\mathrm{K}}$ can also define a \textit{cross-attention}, where keys are obtained by linear transformations of the enriched vision input $\mathbf{X}^{\prime}$ updated periodically and queries - of the enriched vision input updated at each step. This leads to additional computational gains for IAP-rank, as we explain in Sec. \ref{sec:theory}.

\subsection{Image Transformation via IAP Attention - Implicit Attention}
\label{sec:patctopatch}
Computing attention blocks, as defined in Equation \ref{eq:attention}, is in practice very costly when $L$ is large, since it requires explicit construction of the matrix $\mathbf{A} \in \mathbb{R}^{L \times L}$. This means it is not possible to use small-size patches, even for a moderate-size input image, while high-resolution images are intractable. Standard attention modules are characterized by $\Omega(L^{2})$ space and time complexity, where $L$ is the number of patches. We instead propose to leverage $\mathbf{A}$ indirectly, by applying techniques introduced in \citep{performers} for the class of Transformers called \textit{Performers}. We approximate $\mathbf{A}$ via (random) finite feature maps given by the  mapping $\phi:\mathbb{R}^{d_{QK}} \rightarrow \mathbb{R}^{m}$ for a parameter $m \in \mathbb{N}_{+}$, as:
$
\widehat{\mathbf{A}} =  \mathbf{Q}^{\prime}(\mathbf{K}^{\prime})^{\top}, $
where $\mathbf{Q}^{\prime} \in \mathbb{R}^{L \times m}, \mathbf{K}^{\prime} \in \mathbb{R}^{L \times m}$ are matrices with rows: $\phi(\mathbf{q}_{i})$ and $\phi(\mathbf{k}_{i})$ respectively (see: Fig. \ref{fig:iap-scheme}).
By replacing $\mathbf{A}$ with $\widehat{\mathbf{A}}$ in Equation \ref{eq:attention}, we obtain the following efficient version of IAP-rank:  
\begin{equation}
\widehat{\mathrm{Att}}(\mathbf{V}) = \Xi((\mathbf{P}_{1,L} \mathbf{Q}^{\prime})(\mathbf{K}^{\prime})^{\top},\mathbf{V}),    
\end{equation}
where brackets indicate the order of computations. By disentagling $\mathbf{Q}^{\prime}$ from $\mathbf{K}^{\prime}$, we effectively avoid explicitly calculating attention matrices and compute the input to $\Xi$ in linear (rather than quadratic) in $L$ time and space. 
Similarly, for IAP-trans class we proceed with the computations in one of the following orders (depending on whether $l < d_V$ or $l \geq d_V$):
\begin{equation}
\widehat{\mathrm{Att}}(\mathbf{V}) = \Xi(((\mathbf{P}_{l,L} \mathbf{Q}^{\prime})(\mathbf{K}^{\prime})^{\top})\mathbf{V}) \textrm{  or  }    
\Xi(\mathbf{P}_{l,L} (\mathbf{Q}^{\prime}((\mathbf{K}^{\prime})^{\top}\mathbf{V)))}.
\end{equation}

For top-$l$ selection strategy, we used two kernels $\mathrm{K}$ to define attention. The first models \textit{softmax attention} and is of the form:
$\mathrm{K}(\mathbf{u}, \mathbf{v}) = \mathrm{SM}(\mathbf{x},\mathbf{y})$,
for \textit{softmax kernel} $\mathrm{SM}(\mathbf{x},\mathbf{y})\overset{\mathrm{def}}{=}\exp(\mathbf{x}^{\top}\mathbf{y})$ and:
$
\mathbf{x}=d^{-\frac{1}{4}}_{\mathrm{QK}}\mathbf{u}, \mathbf{y}=d^{-\frac{1}{4}}_{\mathrm{QK}}\mathbf{v},
\textrm{ or } 
\mathbf{x}=d^{\frac{1}{4}}_{\mathrm{QK}}\frac{\mathbf{u}}{\|\mathbf{u}\|_{2}}, \mathbf{y}=d^{\frac{1}{4}}_{\mathrm{QK}}\frac{\mathbf{v}}{\|\mathbf{v}\|_{2}},
$
where we call the latter: \textit{normalized query/key} version. The second one is the so-called \textit{Performer-ReLU} variant, given by: $\mathrm{K}(\mathbf{u},\mathbf{v})=\mathrm{ReLU}(\mathbf{u})^{\top}\mathrm{ReLU}(\mathbf{v})$, with $\mathrm{ReLU}$ applied element-wise.
The latter kernel has a trivial corresponding deterministic mapping $\phi$. The one for the former can be obtained from the mapping $\phi$ for the softmax kernel $\mathrm{SM}$. In \citep{performers}, the following map defining random features for the softmax kernel is proposed:
\begin{equation}
\label{positive}
\phi^{m}_{\mathrm{exp}}(\mathbf{z}) = \frac{\Lambda(\mathbf{z})}{\sqrt{m}}
(\exp(\omega_{1}^{\top}\mathbf{z}),...,
\exp(\omega_{m}^{\top}\mathbf{z}))
\end{equation}
for $\Lambda(\mathbf{z})=\exp(-\frac{\|\mathbf{z}\|^{2}}{2})$
and the block-orthogonal ensemble of Gaussian vectors $\{\omega_{1},...,\omega_{m}\}$ with marginal distributions $\mathcal{N}(0,\mathbf{I}_{d_{QK}})$. An alternative more standard map is of the form:
\begin{equation}
\label{trig}
\phi^{m}_{\mathrm{trig}}(\mathbf{z}) = \frac{\Lambda^{-1}(\mathbf{z})}{\sqrt{m}}
(\sin(\omega_{1}^{\top}\mathbf{z}),\cos(\omega_{1}^{\top}\mathbf{z})...,
\sin(\omega_{m}^{\top}\mathbf{z}),\cos(\omega_{1}^{\top}\mathbf{z})).
\end{equation}

\vspace{-3mm}
\paragraph{Strengths \& weaknesses of different kernels/maps $\phi$:} Mapping $\phi$ for Performer-ReLU is fastest to compute (see: inference times in Fig. \ref{fig:intro}), yet softmax attention is in general more expressive. In our experiments for certain benchmarks (e.g. $\mathrm{step}$-$\mathrm{stones}$ task from Sec. \ref{sec:exp}) other attention kernels did not work. Given softmax attention, trigonometric features from Eq. \ref{trig} provide: uniform convergence and (combined with normalized queries/keys) additional computational gains in IAP-rank as compared to positive random features from Eq. \ref{positive}, yet they cannot be applied together with softmax sampling.

%% file: theory.tex
\vspace{-3mm}
\section{The Theory of Implicit Attention for Pixels}
\label{sec:theory}
\vspace{-2mm}
We focus here on the IAP-rank class, giving in particular some of the first results connecting linear attention algorithms with the rich theory of search in the dot-product space (guarantees for IAP-trans follow from theory presented in \citep{performers}). We start with the following definition and lemma:

\begin{definition}
We say that ranking $\mathbf{r}^{\prime}$ approximating $\mathbf{r}$ is $\epsilon$-approximate if $\|\mathbf{r}-\mathbf{r}^{\prime}\|_{\infty} \leq \epsilon$.
We say that an ensemble of $l$ selected patches is $\epsilon$-approximate (with respect to $\mathbf{r}$) if they have top $l$ scores in some $\epsilon$-approximate ranking $\mathbf{r}^{\prime}$.
\end{definition}

\begin{lemma}
\label{first_lemma}
Top $l$ patches computed by IAP-rank (with default IAP-rank setitng: $\mathbf{P}_{1,L}=[1,...,1]$) correspond to
largest $l$ values of $\mathbf{z}^{\top} \phi(\mathbf{k}_{i})$ for $\mathbf{z} = \sum_{j=1}^{L}\phi(\mathbf{q}_{j})$. Furthermore, if trigonometric features from Eq. \ref{trig} are used and queries/keys are normalized, those top $l$ patches correspond also to smallest $l$ values of $\theta_{\mathbf{z}, \phi(\mathbf{k}_{i})}$, where $\theta_{\mathbf{x},\mathbf{y}} \in [0, \pi]$ stands for an angle between $\mathbf{x}$ and $\mathbf{y}$.
\end{lemma}

The above lemma is a gateway to replacing time complexity $\Omega(Lm)$ of finding the $l$ top patches (given computed: $(\phi(\mathbf{q}_{i}),\phi(\mathbf{k}_{i}))_{i=1,...,L}$) with time complexity $\Gamma^{l}_{\mathbf{z}}(\phi(\mathbf{k}_{1}),...,\phi(\mathbf{k}_{L}))$ of any algorithm finding largest $l$ dot-products $\mathbf{z}^{\top}\phi(\mathbf{k}_{i})$.
Computational gains (also for the softmax sampling version of the algorithm) come in the cross-attention setting, where keys are updated with lower frequency than queries and the problem effectively reduces to querying a database for top $l$ dot-products, where a \textit{key-database} (a set of $\phi$-transformed keys) is recomputed less frequently. This is demonstrated below. The groundtruth ranking $\mathbf{r}$ is given here by IAP-rank attention with no hashing.
\begin{theorem}
\label{first_theorem}
Consider hashing defined as: $\mathbf{x} \rightarrow \frac{1}{\sqrt{m^{\prime}}} \mathrm{sign}(\mathbf{Gx})$ for Gaussian $\mathbf{G} \in \mathbb{R}^{m^{\prime} \times 2m}$ applied to $(\phi(\mathbf{q}_{i}),\phi(\mathbf{k}_{i}))_{i=1,...,L}$.
For the IAP-rank with cross-attention, trigonometric features, queries/keys normalization and with the above hashing applied prior to choosing top patches, $O(Lm)$ time complexity of ranking is replaced by $O(Lm^{\prime})$ with the one-time cost per key-database update $O(Lmm^{\prime})$. For $m^{\prime} \geq \frac{2}{p\epsilon^{2}}(\log(L)+\log(I))$, the procedure outputs an  $\exp(\sqrt{d_{QK}})\frac{\pi}{2}\epsilon$-approximate ensemble of patches with probability $\geq 1-p$ across $I$ steps. If positive features from Eq. \ref{positive} are applied, approximate softmax sampling (with bias going to $0$ as $m \rightarrow \infty$) can replace the above $O(Lm)$ per step time by $O(\log(L)m)$ with the one-time cost $O(Lm)$ per key-database update. 
\end{theorem}

Our main theoretical result, regarding the quality of IAP-rank approximation of the brute force softmax attention, is given below. The groundtruth ranking $\mathbf{r}$ is given here by brute force attention. 
\begin{theorem}[IAP-rank approximating brute force softmax attention]
\label{second_theorem}
IAP-rank applying trigonometric features from Eq. \ref{trig} with $m \geq \frac{4(d_{QK}+2)(10+\log(\frac{R^{2}\sqrt{d_{QK}}}{p\epsilon^{2}})+\frac{R^{2}}{\sqrt{d_{QK}}})}{\epsilon^{2}}$ random projections and no queries/keys normalization, provides $\epsilon$-accurate ranking with probability $\geq 1-p$ if queries and keys are taken from the ball of radius $R$.
Furthermore, if in the queries/keys normalization setting, the $L_{\infty}$ norms of the columns of $\mathbf{A}_{\mathrm{K}}$ corresponding to tokens from a given set $\mathcal{P}$ are upper-bounded by $\exp(\sqrt{d_{QK}}\cos(\pi-\alpha))$ for some $\alpha \in [0, \pi]$, then their corresponding approximate scores given by IAP-rank with positive features from Eq. \ref{positive} are at most $\epsilon$ with probability $\geq 1 -p$ over $I$ steps if:
\begin{equation}
m \geq \frac{|\mathcal{P}|^{2}I}{p\epsilon^{2}}\exp(8\sqrt{d_{QK}}\sin^{2}(\frac{\alpha}{2})-2\sqrt{d_{QK}})(1-\exp(-4\sqrt{d_{QK}}\sin^{2}(\frac{\alpha}{2}))).    
\end{equation}
\end{theorem}

Here $|\mathcal{P}|$ stands for the size of $\mathcal{P}$. In the above theorem, set $\mathcal{P}$ for small $\alpha$ corresponds to unimportant patches and the theorem says that those patches with high probability get assigned low scores by IAP-rank with positive features and small (in comparison to $L$) number of random projections $m$.

%% file: experiments.tex
\vspace{-2mm}
\section{Experiments}
\label{sec:exp}
\vspace{-3mm}

In this section we show that IAP achieves strong results in RL, also for high-resolution input data, by testing it exhaustively on a large set of problems ranging from challenging large scale vision tasks with distractions to difficult locomotion and navigation tasks involving quadruped robots: obstacle avoidance, navigation and walking on uneven terrains. We demonstrate that the method is scalable enough to model even pixel-to-pixel attention and that smaller patches are particularly effective in preventing observational overfitting in the presence of distractions. All controllers are trained with ES methods \cite{ES}. We encourage Reader to check attached supplement for the videos of trained policies and comments on: what IAP-agents attend to as well as styles of IAP-policies for locomotion.

In the introduction we have already demonstrated (see: Fig. \ref{fig:intro}) that in practice one does not need many random features to learn good quality RL controllers
which implies faster training and inference. In those experiments we used observations resized to (100 x 100), similar to the (96 x 96) sizes used for $\mathrm{CarRacing}$ and $\mathrm{DoomTakeCover}$ in \cite{yujintang}, patches of size $4$ and selected the top $l=5$ patches.
\vspace{-3mm}
\subsection{Distracting Control Suite}
\label{sec:distracting_controlsuite}
\vspace{-3mm}
We then apply our method to a modified version of the DM control suite called the \textit{Distracting Control Suite} \cite{distracting}, with backgrounds of the normal DM Control Suite's observations replaced with random images and viewed through random camera angles as shown in Fig. \ref{fig:dcs} in the Appendix.
We used IAP-rank with no masking and different kernels and maps $\phi$.

\begin{figure}[h]
\vspace{-1mm}
\centering
    \includegraphics[keepaspectratio, width=0.99\textwidth]{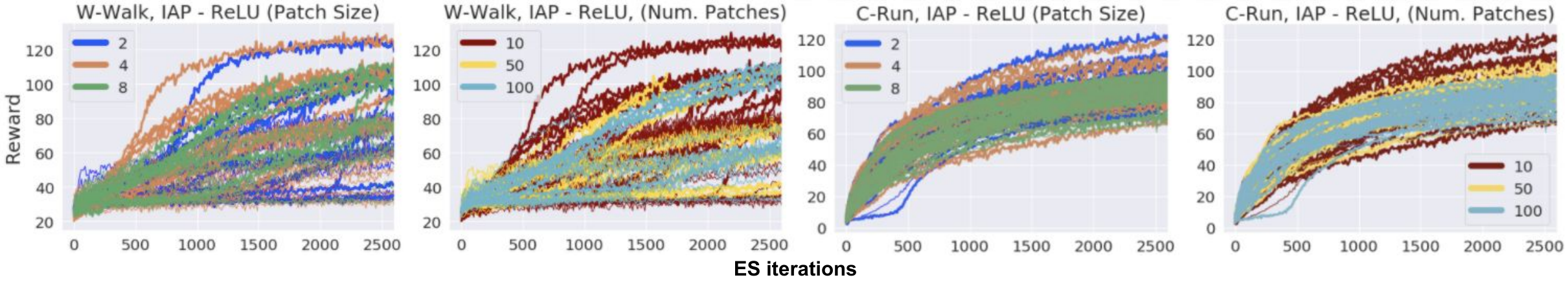}
    \vspace{-3mm}
\caption{\small{A grid-search sweep over patch sizes in $\{2, 4, 8\}$, embedding dimensions in $\{4, 16, 64\}$, and number of patches $l \in \{10, 50, 100\}$. We see that generally, smaller patch sizes with lower $l$ improves performance.}}
\label{fig:patch_comparison} 
\vspace{-3mm}
\end{figure}

By default in this benchmark, the native images are of size (240 x 320), substantially larger than (96 x 96) used in \cite{yujintang}, and given that we may also use smaller patch sizes (e.g. size 2 vs the default 7 in \cite{yujintang}), this new benchmark leads to a significantly longer maximum sequence length $L$ (\textbf{19200 vs 529}) for the attention component. In addition, given the particularly small stick-like appearances of most of the agents, a higher percentage of image patches will contain irrelevant background observations that can cause \textit{observational overfitting} \cite{Song2020Observational}, making this task more difficult for vision-based policies.

Our experimental results on the Distracting Control Suite show that more fine-grained patches (lower patch size) with fewer selected patches (lower $l$) improves performance  (Fig. \ref{fig:patch_comparison}). Interestingly, this is contrary to the results found in \cite{yujintang}, which showed that for $\mathrm{CarRacing}$ with YouTube/Noisy backgrounds, decreasing $l$ \textit{reduces} performance as the agent attends to noisier patches. We hypothesize this could be due to many potential reasons (higher parameter count from ES, different benchmarks, bottleneck effects, etc.) but we leave this investigation to future works.

\begin{figure}[h]
\centering
    \includegraphics[keepaspectratio, width=0.49\textwidth]{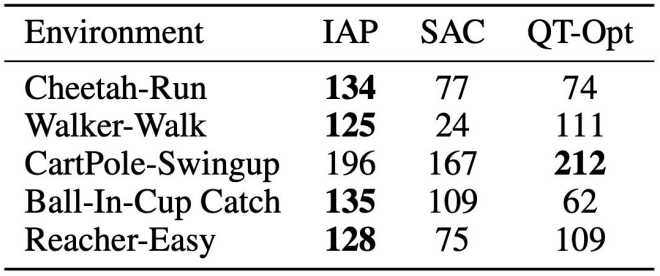}
    \vspace{-3mm}
\caption{\small{We use the \textbf{static} setting on the \textbf{medium difficulty} benchmark found in \cite{distracting}. We include reported results from the paper for SAC and QT-Opt. For IAP, we report the final reward for the fastest convergent method. }}
\label{fig:method_comparison} 
\vspace{-3mm}
\end{figure}

We thus use patch sizes of 2 with $l = 10$ patches. Furthermore, we compare our algorithm with standard ConvNets trained with SAC \cite{sac-v2} and QT-Opt \cite{qt_opt} in Fig. \ref{fig:method_comparison} and find that we are consistently competitive or outperform those methods.
\vspace{-3mm}
\subsection{Visual Locomotion and Navigation Tasks with Quadruped Robots}
\label{sec:quadruped_locomotion}
We set up three different types of vision-based robotic locomotion tasks to test IAP as follows. 


\textbf{Navigating in obstacle course:}
For this task, we use a simulated quadruped with $12$ degrees of freedom ($3$ per leg). The robot starts from the origin on a raised platform and a series of walls lies ahead of it. The robot perceives through a first-person RGB-camera view, looking straight ahead. 
The robot's task is to progress forward as fast as possible. 
It needs to learn to steer in order to avoid collisions with the walls and falling off the edge. The reward for each timestep is the capped ($v_{cap}$) velocity of the robot along the $x$ direction (see: Section \ref{locomotion_details}).

\begin{figure*}[h]
\centering
\includegraphics[keepaspectratio, width=0.99\textwidth]{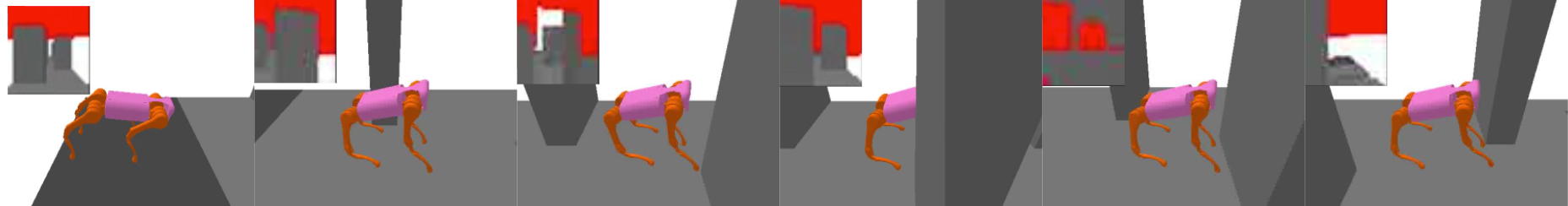} \\
\includegraphics[keepaspectratio, width=0.99\textwidth]{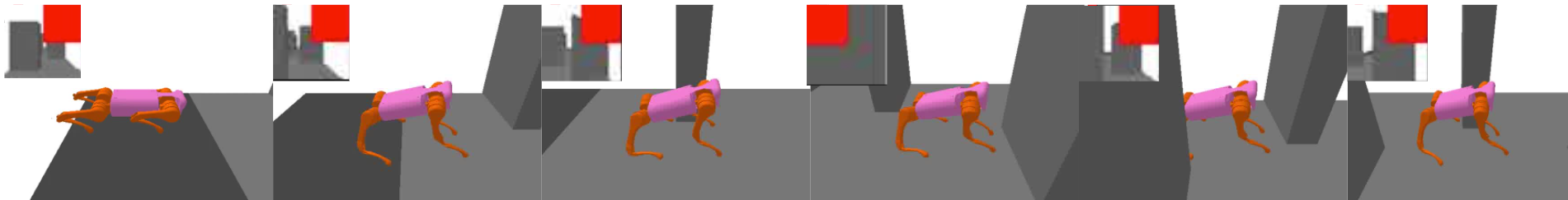}
\vspace{-3mm}
\caption{\small{IAP policies with patch size $1$ (pixel-to-pixel attention, top row) and patch size $16$ (bottom row) for the navigation in obstacle course task. A series of image frames along the episode length are shown. On the top-left corner of the images, the input camera image is attached. The red part of the camera image is the area selected by self-attention. In case of patch size $1$, we can see that the policy finely detects the boundaries of the obstacles. For patch size $16$, only a single patch is selected which covers one fourth of the whole camera image. The policy identifies general walking direction but fine-grained visual information is lost.}}
\label{fig:patch_viz}
\vspace{-2.5mm}
\end{figure*}

Given the complexity of the task, we use a hierarchical structure for our policies introduced in \cite{Jain2019HierarchicalRL}. 
In this setup, the policy is split into two hierarchical levels - high and low. The high level processes a $32 \times 32 \times 3$ RGB camera image using IAP-rank without masking and outputs a latent command for the low level. IAP uses deterministic ReLU features. The high level also outputs a scalar duration for which its execution is stopped, while the low level runs at every control timestep. The low level is a linear neural network which controls the robot leg movements. Policies with patch size $1$ and patch size $16$ are visualized in Fig. \ref{fig:patch_viz}.

\textbf{Locomotion on uneven terrains:} We use the quadruped from Unitree called Laikago~\citep{laikagounitree} for learning locomotion on uneven terrains. It has $12$ actuated joints, $3$ per leg. It has to walk forward on a variety of randomized uneven terrains that requires careful foot placement planning based on visual feedback. It perceives the ground through $2$ depth cameras attached to its body, one on the front and other on the belly facing downwards. IAP processes $32 \times 24$ depth images from these $2$ cameras. We test the vision policy to walk on the following types of randomized terrain:
\vspace{-3mm}
\begin{enumerate}
    \item $\mathrm{Step}$-$\mathrm{stones}$: The ground is made of a series of stepstones with gaps in between. The step stones widths are fixed at $50$~cm, the lengths are between $[50,80]$~cm in length, and the gap size between adjacent stones are between $[10,20]$~cm.
    \item $\mathrm{Grid}$: The ground is a grid of small square step-stones of size $15 \times 15 ~\mathrm{cm^2}$. They are separated by $[13, 17]$~cm from each other in both $x$ and $y$ directions. At the beginning of each episode, we also randomly rotate entire grid by an angle sampled in $[-0.1, 0.1]$~radians.
    \item $\mathrm{Stairs}$: The robot needs to climb up a flight of stairs. The depth of each stair is uniformly randomized in the range $[25, 33]$~cm and the height is in the range $[16, 19]$~cm. 
\end{enumerate}
\vspace{-3mm}

\begin{figure*}[h]
\centering
\includegraphics[keepaspectratio, width=0.99\textwidth]{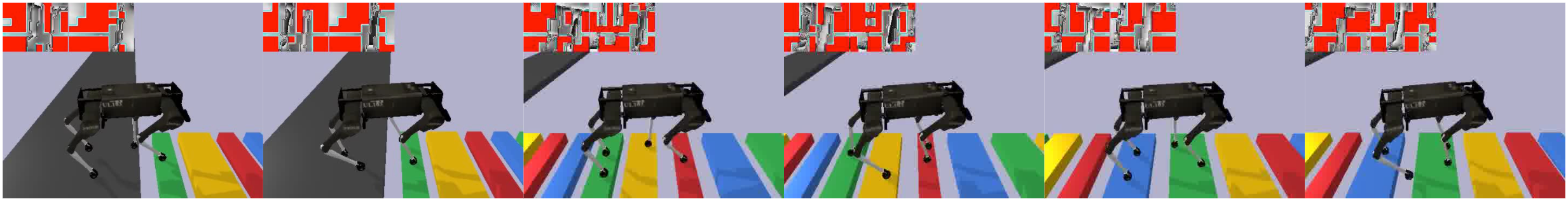}
\includegraphics[keepaspectratio, width=0.99\textwidth]{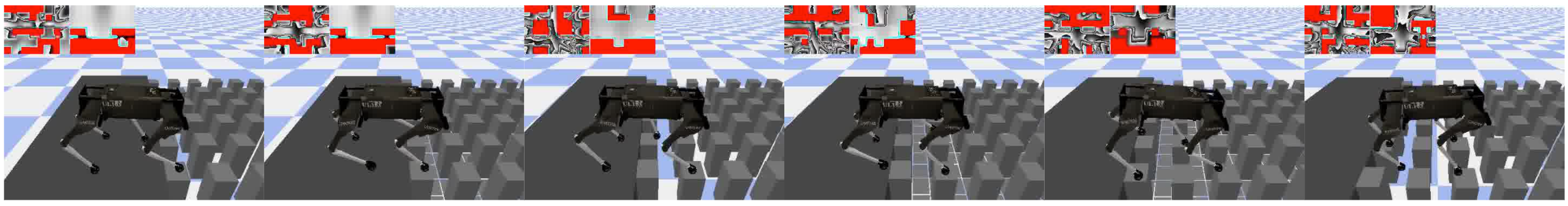}
\includegraphics[keepaspectratio, width=0.99\textwidth]{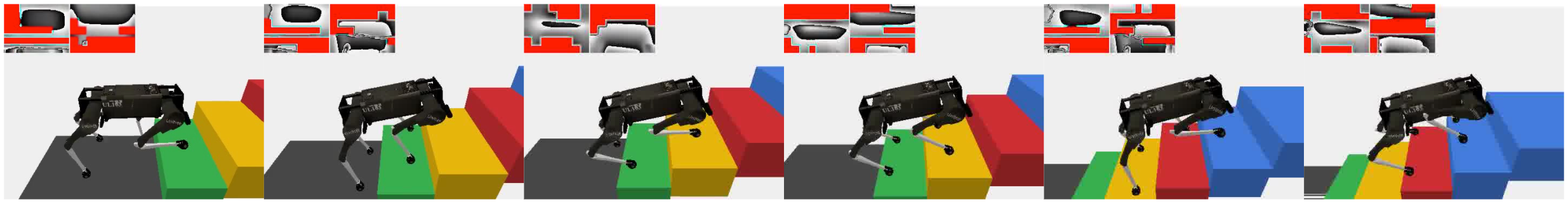}
\vspace{-3mm}
\caption{\small{IAP policies walking on uneven terrains: step-stone (top), grid (middle) and stairs (bottom).}}
\label{fig:mpc}
\vspace{-2.5mm}
\end{figure*}

\begin{figure*}[h]
\centering
\includegraphics[keepaspectratio, width=0.99\textwidth]{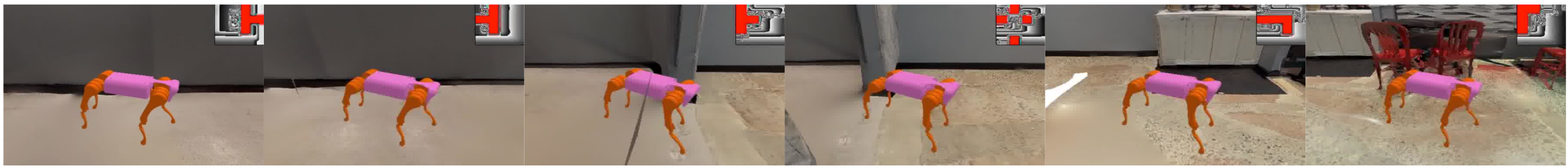}
\vspace{-3mm}
\caption{\small{IAP policy navigating photo-realistic indoor Gibson environment.}}
\label{fig:gibson}
\vspace{-2.5mm}
\end{figure*}

\begin{figure*}
\vspace{-3mm}
\centering
    \includegraphics[keepaspectratio, width=0.48\textwidth]{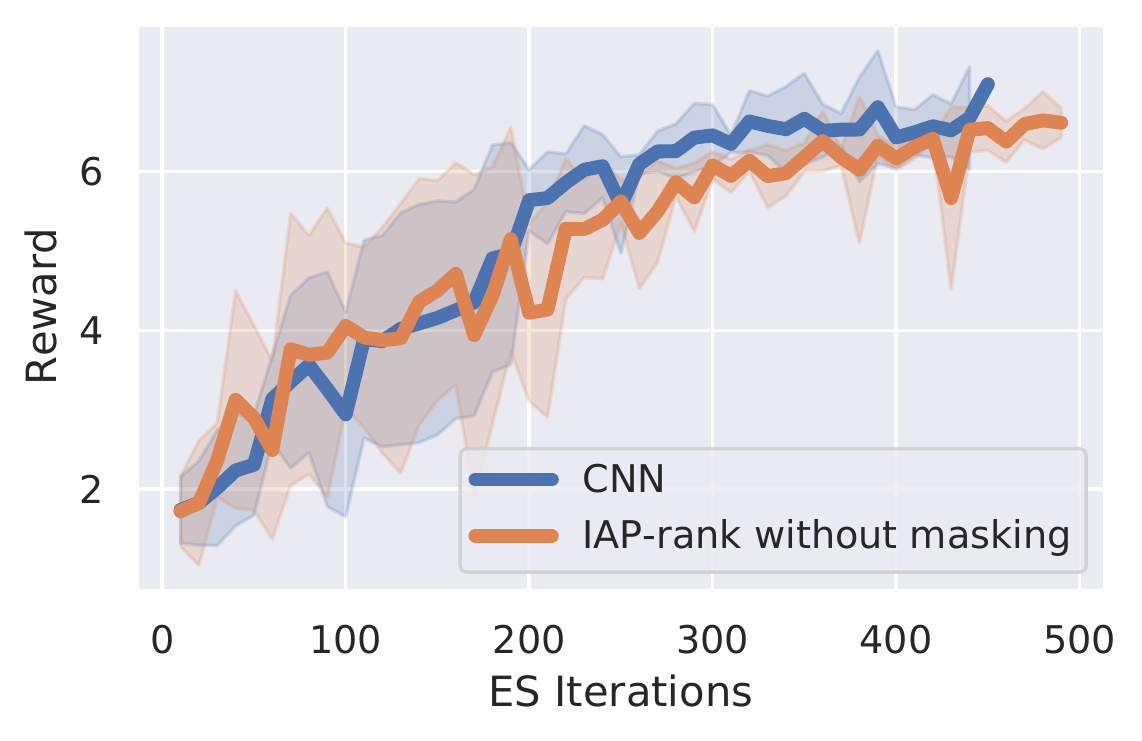}
    \includegraphics[keepaspectratio, width=0.48\textwidth]{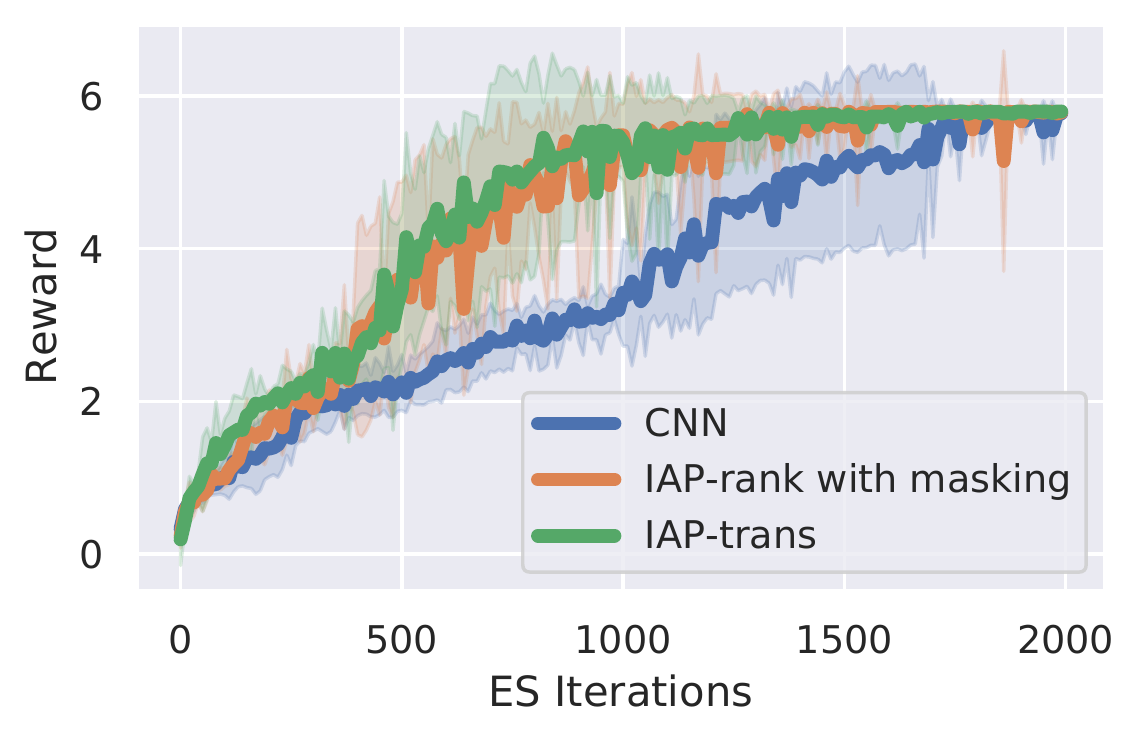}
\vspace{-3mm}
\caption{\small{IAP- versus CNN-policies on obstacle course navigation (Left) and step-stones env (Right). IAP is competitive with or outperforms its CNN counterparts, requiring substantially fewer parameters. For obstacle course navigation, ReLU features suffice. For the step-stones env, we applied more expressive softmax features.}}
\label{fig:cnn_vs_att} 
\vspace{-0.2cm}
\end{figure*}

As before, we use a hierarchical setup for solving these tasks. The high level uses IAP-rank with masking or IAP-trans to process camera images and output the desired foot placement position. In these tasks, we use softmax kernel features in IAP. The low level employs a position-based swing leg controller, and an model predictive control (MPC) based stance leg controller~\cite{bledt2018cheetah}, to achieve the foot placement decided by high level. Performance of learned policies is shown in Fig. \ref{fig:mpc}.

\textbf{Navigating in photo-realistic Gibson environments:}
We also trained interpretable IAP policies from scratch for locomotion and navigation in simulated 3D-spaces with realistic visuals from the Gibson dataset~\cite{xiazamirhe2018gibsonenv}. A learned policy is shown in Fig. \ref{fig:gibson}. 


Training curves for the CNN and IAP policies are shown in Fig. \ref{fig:cnn_vs_att}. We observe similar task performance for IAP-rank without masking and CNN in the obstacle course navigation task. IAP-rank with masking and IAP-trans train faster than CNN in the task of locomotion on uneven terrain (stepstones). More importantly, IAP-rank without masking (selecting top-l patches) has \textbf{8.4} times fewer parameters and IAP-rank with masking as well as IAP-trans \textbf{4.2} times fewer parameters than CNN-policies.
We trained IAP with different values for the patch size parameter ($1$, $4$, $8$ and $16$). Maximum episode return is achieved by patch size $1$ - a setting with the largest number of patches (pixel-to-pixel attention) (see: Section \ref{sec:patch_ablations}). 
Videos of IAP performance can be viewed \href{https://sites.google.com/view/implicitattention}{here\footnote{\href{https://sites.google.com/view/implicitattention}{https://sites.google.com/view/implicitattention}}}.

%% file: conclusion.tex
\vspace{-3mm}
\section{Conclusion}
\vspace{-3mm}
In this paper, we significantly expanded the capabilities of methods using attention in RL. We are the first to show that efficient attention mechanisms, which have recently demonstrated impressive results for Transformers, can be used for RL policies, in what we call \emph{Implicit Attention for Pixels} or IAP. In a series of experiments, we showed that IAP scales to higher-resolution images and emulates much finer-grain attention than what was previously possible, improving generalization in challenging vision-based RL involving quadruped robots and the recently introduced Distracting Control Suite.

%% file: appendix.tex
\section{APPENDIX: Unlocking Pixels for Reinforcement Learning via Implicit Attention}

Due to social distance policies implemented because of the pandemic, all our experiments with quadruped robots for locomotion/navigation were conducted in the simulator which was accurately reflecting real on-robot setup (see in particular: comments below regarding the Laikago robot). We provide details regarding this setup in the following paragraphs. That simulator was used in several previous papers on quadruped locomotion to provide good quality sim policies that were successfully transferable to real hardware.

\subsection{Extra Figures}
\begin{figure}[h]
\centering
\includegraphics[keepaspectratio, width=0.99\textwidth]{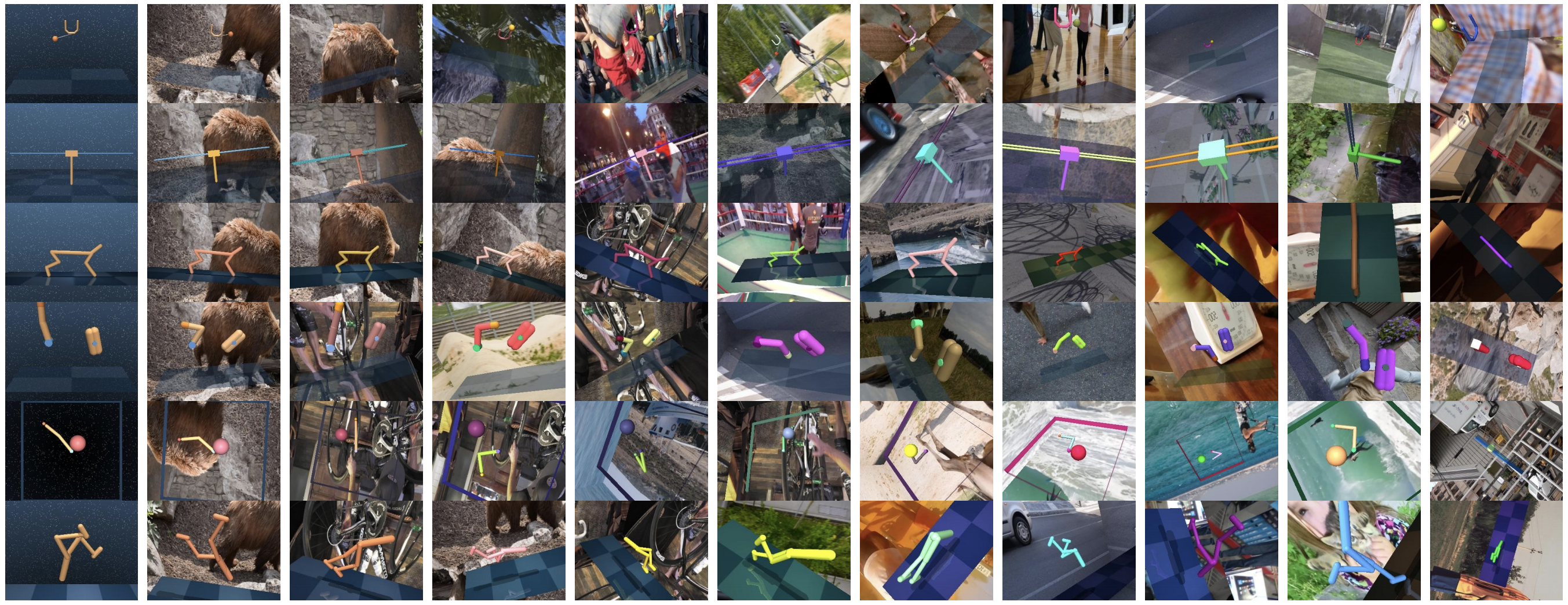}
\caption{\small{Examples of Distracting Control Suite \cite{distracting} tasks with distractions in the background that need to be automatically filtered out to learn a successful controller. Image resolutions are substantially larger than for most other vision-based benchmarks for RL considered before. Code can be found at \url{https://github.com/google-research/google-research/tree/master/distracting_control}.}}
\label{fig:dcs} 

\end{figure}

\subsection{Quadruped Locomotion Experiments}
\label{locomotion_details}
We provide here more details regarding an experimental setup for the quadruped locomotion tasks. 

Our simulated robot is similar in size, actuator performance, and range of motion to the MIT Mini Cheetah~\cite{minicheetah} ($9$ kg) and Unitree A1\footnote{\href{https://www.unitree.com/products/a1/}{https://www.unitree.com/products/a1/}} ($12$ kg)  robots. Robot leg movements are generated using a trajectory generator, based on the \emph{Policies Modulating Trajectory Generators} (PMTG) architecture,  which has shown success at learning diverse primitive behaviors for quadruped robots~\cite{iscen2018policies}. The latent command from the high level, IMU sensor observations, motor angles, and current PMTG state is fed to low level neural network which outputs the residual motor commands and PMTG parameters at every timestep. 

We use the Unitree A1's URDF description\footnote{\href{https://github.com/unitreerobotics}{https://github.com/unitreerobotics}}, which is available in the PyBullet simulator~\cite{pybulletcoumans}.
The swing and extension of each leg is controlled by a PD position controller. 

The reward function is specified as the capped ($v_{cap}$) velocity of the robot along the x direction:

\begin{align}
f_{v_{cap}}(r) &= \max(-v_{cap},\min(r, v_{cap}))\\
    r_{cc}(t) &= f_{v_{cap}}(x(t) - x(t-1)).
\end{align}

\subsubsection{Ablations over Patch Sizes}
\label{sec:patch_ablations}
In Table 1 we present detailed results regarding ablations with patch sizes and stride lengths for the navigation with obstacle avoidance task. As explained in the main body, highest rewards are achieved for the pixel-to-pixel attention corresponding to patch size $1$.
\begin{table}[H]
\centering
    \caption{\small{Ablation with patch sizes and stride length for the navigation with obstacle avoidance task.}}
    \label{tab:loco_abl}
    \scalebox{0.85}{
    \begin{tabular}{ccr}
    \toprule
    \multicolumn{1}{l}{Patch Size} & \multicolumn{1}{l}{Stride Length} & \multicolumn{1}{l}{Maximum Reward} \\ \midrule
    1                                         & 1                                           & \textbf{8.0}                            \\ \hline
    \multirow{2}{*}{4}                        & 2                                           & 6.9                                     \\ \cline{2-3} 
                                              & 4                                           & 7.5                                     \\ \hline
    \multirow{2}{*}{8}                        & 4                                           & 6.3                                     \\ \cline{2-3} 
                                              & 8                                           & 7.5                                     \\ \hline
    \multirow{2}{*}{16}                       & 8                                           & 6.6                                     \\ \cline{2-3} 
                                              & 16                                          & 7.6                                     \\ \bottomrule
    \end{tabular}}
\end{table}

\subsubsection{Policy Behavior}
\label{sec:iap_behavior}
We present videos for locomotion behavior of IAP policies in attached supplementary material. In all simulation videos, the input camera images are attached on the top left corner of the video. The red part of the camera image is the area selected by self-attention. 

Videos \texttt{p1dt\_fast.mp4} and \texttt{p16dt\_fast.mp4} show obstacle avoidance policies. In case of patch size = 1 (\texttt{p1dt\_fast.mp4}), we can see that the policy finely detects the boundaries of the obstacles which helps in navigation. For patch size = 16 (\texttt{p16dt\_fast.mp4}), only a single patch is selected which covers one fourth of the whole camera image. The policy identifies general walking direction but fine-grained visual information is lost.

\begin{figure}[h]
\centering
\includegraphics[keepaspectratio, width=0.49\textwidth]{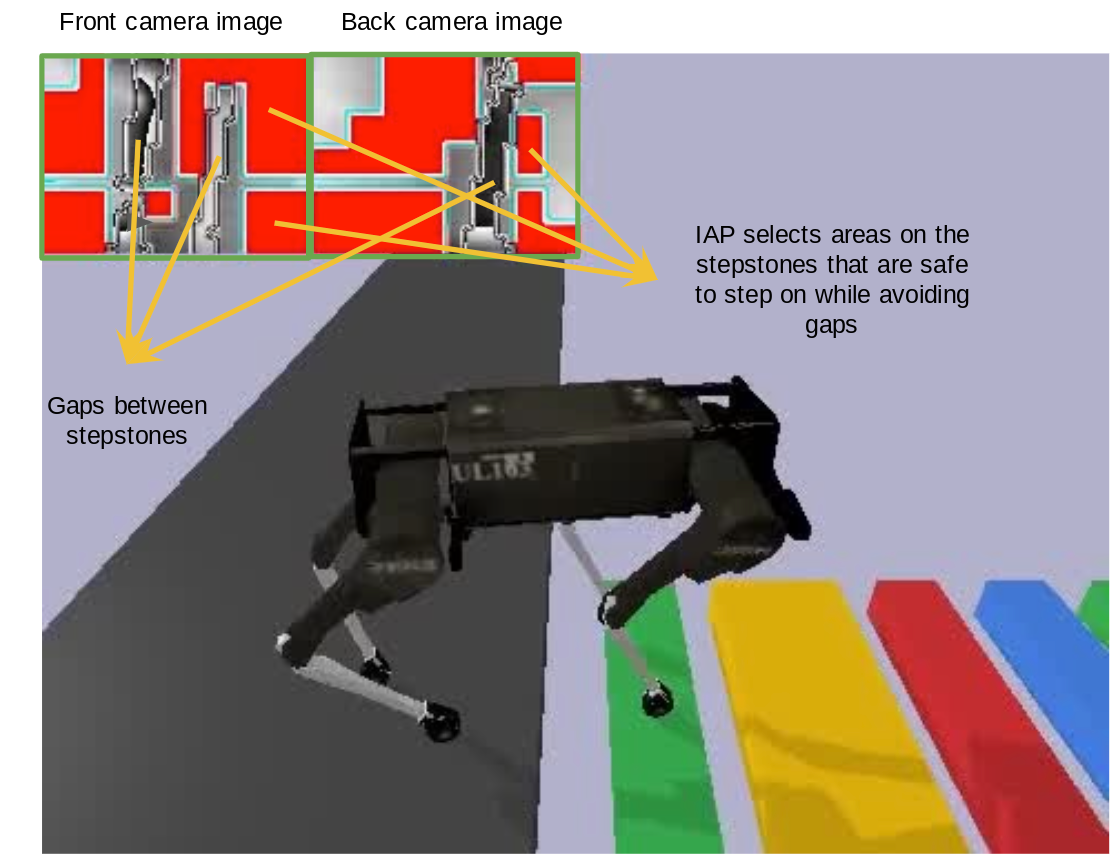}
\caption{\small{Interpreting IAP policy behavior while walking on step-stones.}}
\label{fig:iap_vis} 
\end{figure}

Video \texttt{mpc\_ss.mp4} shows walking behaviour on uneven step-stones. Notice that IAP selects areas corresponding to the step-stones while ignoring the gaps. These are the places which are safe to step on (see Fig.~\ref{fig:iap_vis}). Based on this selection, the policy picks favorable foot placement location and the MPC based low-level controller adjusts step length to reach desired position, thus avoiding falling in the gap. An interesting observation regarding this policy is that the robot consistently uses its front right to cross the gap first.

Laikago climbs upstairs in \texttt{mpc\_stairs.mp4}. IAP successfully climbs up a flight of stairs by selecting safe horizontal area on the next step.

Video \texttt{mpc\_poles.mp4} shows Laikago walking on a grid of small step-stones. IAP learns to carefully walk on the grid by attending to the stones. 

Video \texttt{video\_gibson.mp4} shows a walking policy in photo-realistic indoor environment. The robot successfully passes through a narrow gate and the policy is focusing on obstacles and pathways in the scene.

All the videos are also available here: \href{https://sites.google.com/view/implicitattention}{https://sites.google.com/view/implicitattention}.

\subsection{Proofs}

\subsubsection{Proof of Lemma \ref{first_lemma}}

\begin{proof}
Consider expression $\mathbf{r} = \frac{1}{L}\mathbf{P}_{1,L}\mathbf{A}_{\mathrm{K}}$ in the IAP-rank algorithm defining the vector of scores.
Note that we have: 
\begin{equation}
\mathbf{r} = \frac{1}{L}([1,...1]\mathbf{Q}^{\top})(\mathbf{K}^{\prime})^{\top} = \frac{1}{L}\mathbf{z}^{\top}(\mathbf{K}^{\prime})^{\top}=
\frac{1}{L}(\mathbf{z}^{\top}\phi(\mathbf{q}_{1}),...,\mathbf{z}^{\top}\phi(\mathbf{q}_{L}))^{\top},    
\end{equation}
where $\mathbf{z} \overset{\mathrm{def}}{=} \sum_{j=1}^{L} \phi(\mathbf{q}_{j})$. Therefore ith score is given as: $r_{i} = \frac{1}{L}\mathbf{z}^{\top}\phi(\mathbf{k}_{i})$ and thus choosing top $l$ patches is equivalent to choosing $l$ patches with the largest value of the corresponding dot-product $\mathbf{z}^{\top}\phi(\mathbf{k}_{i})$. That proves the first part of the lemma. Now assume that all keys are normalized and trigonometric features are used, i.e.
\begin{equation}
\phi(\mathbf{k}_{i}) = \phi^{m}_{\mathrm{trig}}\left(d_{QK}^{\frac{1}{4}}\frac{\mathbf{k}_{i}}{\|\mathbf{k}_{i}\|_{2}}\right)    
\end{equation}
for 
\begin{equation}
\phi^{m}_{\mathrm{trig}}(\mathbf{z}) \overset{\mathrm{def}}{=}
\frac{\Lambda^{-1}(\mathbf{z})}{\sqrt{m}}(\sin(\omega_{1}^{\top}\mathbf{z}),
\cos(\omega_{1}^{\top}\mathbf{z}),
...,
\sin(\omega_{m}^{\top}\mathbf{z}),
\cos(\omega_{m}^{\top}\mathbf{z})
)^{\top}
\end{equation}
for some $m \in \mathbb{N}_{+}$, Gaussian vectors $\omega_{1},...,\omega_{m} \overset{\mathrm{iid}}{\sim} \mathcal{N}(0,\mathbf{I}_{d_{QK}})$ and $\Lambda(\mathbf{z}) \overset{\mathrm{def}}{=} \exp(-\frac{\|\mathbf{z}\|^{2}}{2})$.
Therefore we have: $\|\phi^{m}_{\mathrm{trig}}(\mathbf{z})\|_{2}=\exp(\frac{\|\mathbf{z}\|^{2}}{2})$ and thus:
\begin{equation}
\|\phi(\mathbf{k}_{i})\|_{2} = \exp\left(\frac{\sqrt{d_{QK}}}{2}\right)    
\end{equation}

Therefore we have:
\begin{equation}
\label{cos_eq}
r_{i} = \|\mathbf{z}\|_{2}\|\phi(\mathbf{k}_{i})\|_{2}
\cos(\theta_{\mathbf{z},\phi(\mathbf{k}_{i})})
= \|\mathbf{z}\|_{2}\exp\left(\frac{\sqrt{d_{QK}}}{2}\right)
\cos(\theta_{\mathbf{z},\phi(\mathbf{k}_{i})})
\end{equation}
Thus top $l$ scores correspond to smallest $l$ values of
$\cos(\theta_{\mathbf{z},\phi(\mathbf{k}_{i})}) \in [0, \pi]$.
That completes the proof.
\end{proof}

\subsubsection{Proof of Theorem \ref{first_theorem}}
\begin{proof}
We start with the part of the theorem regarding trigonometric features.
Consider transformation: $\psi(\mathbf{z}) = \frac{1}{\sqrt{m}}\mathrm{sign}(\mathbf{Gz})$ for the Gaussian matrix $\mathbf{G} \in \mathbb{R}^{m^{\prime} \times 2m}$ with entries taken independently at random from $\mathcal{N}(0,1)$ and $\mathbf{z} \in \mathbb{R}^{2m}$. Note first that for $\mathbf{z}_{1},\mathbf{z}_{2} \in \mathbb{R}^{d}$ for $d=2m$ the following holds:
\begin{equation}
\label{ang_kernel}
\mathbb{E}[\psi(\mathbf{z}_{1})^{\top}\psi(\mathbf{z}_{2})] = 
1 - \frac{2\theta_{\mathbf{z}_{1},\mathbf{z}_{2}}}{\pi},
\end{equation}
where $\theta_{\mathbf{x},\mathbf{y}} \in [0, \pi]$ stands for an angle between $\mathbf{x}$ and $\mathbf{y}$.
To prove it, let us define: 
\begin{equation}
X_{i} = \psi(\mathbf{z}_{1})(i)\psi(\mathbf{z}_{2})(i)    
\end{equation}
It suffices to show that: $\mathbb{E}[X_{i}] = \frac{1}{m^{\prime}} (1 - \frac{2\theta_{\mathbf{z}_{1},\mathbf{z}_{2}}}{\pi})$.
Note that we have: 
\begin{equation}
X_{i} = \frac{1}{m^{\prime}}\mathrm{sign}(\omega_{i}^{\top}\mathbf{z}_{1})
\mathrm{sign}(\omega_{i}^{\top}\mathbf{z}_{2}) = \frac{1}{m^{\prime}}\mathrm{sign}((\omega^{\mathrm{proj}}_{i}(\mathbf{z}_{1},\mathbf{z}_{2}))^{\top}\mathbf{z}_{1})
\mathrm{sign}((\omega^{\mathrm{proj}}(\mathbf{z}_{1},\mathbf{z}_{2}))_{i}^{\top}\mathbf{z}_{2}),    
\end{equation}
where $\omega_{i}$ is the vectorized ith row of $\mathbf{G}$ and $\omega^{\mathrm{proj}}_{i}(\mathbf{z}_{1},\mathbf{z}_{2})$ is its projection into $2$-dimensional linear space spanned by $\{\mathbf{z}_{1},\mathbf{z}_{2}\}$.
Define $\mathcal{E}_{i}$ as an event that $\omega^{\mathrm{proj}}_{i}(\mathbf{z}_{1},\mathbf{z}_{2})^{\top}\mathbf{z}_{1}$ and $\omega^{\mathrm{proj}}_{i}(\mathbf{z}_{1},\mathbf{z}_{2})^{\top}\mathbf{z}_{2}$ have different signs. From the definition of $X_{i}$ we have:
\begin{equation}
\mathbb{E}[X_{i}] = \frac{1}{m^{\prime}}((-1)\cdot \mathbb{P}[\mathcal{E}_{i}] +  (1-\mathbb{P}[\mathcal{E}_{i}]))= \frac{1}{m^{\prime}}(1-2\mathbb{P}[\mathcal{E}_{i}])    
\end{equation}
Thus it suffices to show that: $\mathbb{P}[\mathcal{E}_{i}]) = \frac{\theta_{\mathbf{z}_{1},\mathbf{z}_{2}}}{\pi}$, but this is a direct implication of the fact that the distribution of 
$\omega^{\mathrm{proj}}_{i}(\mathbf{z}_{1},\mathbf{z}_{2})$ is isotropic in the $2$-dimensional space spanned by $\{\mathbf{z}_{1},\mathbf{z}_{2}\}$. That proves Equation \ref{ang_kernel}.

Notice that:
\begin{align}
\begin{split}
\mathbb{P}\left[\left|\psi(\mathbf{z}_{1})^{\top}\psi(\mathbf{z}_{2}) - (1-\frac{2\theta_{\mathbf{z}_{1},\mathbf{z}_{2}}}{\pi})\right| > \epsilon\right] = \mathbb{P}[|\sum_{i=1}^{L}X_{i}-\mathbb{E}[\sum_{i=1}^{L}X_{i}]|>\epsilon] \leq 2\exp(-\frac{m\epsilon^{2}}{2}), 
\end{split}
\end{align}
where the last inequality follows from Azuma's Inequality.
Denote by $\mathcal{E}$ an event that for all $i=1,...,L$ the following holds: $\left|\psi(\mathbf{z})^{\top}\psi(\phi(\mathbf{k}_{i})) - (1-\frac{2\theta_{\mathbf{z},\phi(\mathbf{k}_{i})}}{\pi})\right| \leq \epsilon$ across all $I$ steps. By the union bound, we have:
\begin{equation}
\mathbb{P}[\mathcal{E}] \geq 1-2\exp(-\frac{m\epsilon^{2}}{2})LI \end{equation}
\end{proof}
Note that mapping $\psi$ effectively leads to the approximation of angles $\theta_{\mathbf{z},\phi(\mathbf{k}_{i})}$.
if $\mathcal{E}$ holds, then each of this approximation is within $\frac{\pi}{2}\epsilon$ from the groundtruth angle.
Thus, by Equation \ref{cos_eq} and from the fact that cosine function is $1$-Lipschitz, the corresponding approximate scores
are within $\frac{1}{L}\|\mathbf{z}\|_{2}\exp(\frac{\sqrt{d_{QK}}}{2})\frac{\pi}{2}\epsilon$ error from the original IAP-rank scores. Since for every $i$ we have: $\|\phi(\mathbf{q}_{i})\|_{2}=\exp(\frac{\sqrt{d_{QK}}}{2})$, the following holds by triangle inequality:
\begin{equation}
\|\mathbf{z}\|_{2} = \|\sum_{i=1}^{L}\phi(\mathbf{k}_{i})\|_{2} \leq \sum_{i=1}^{L}\|\phi(\mathbf{k}_{i})\|_{2} = L\exp(\frac{\sqrt{d_{QK}}}{2})    
\end{equation}
Thus the approximate scores are within $\exp(\frac{\sqrt{d_{QK}}}{2})\frac{\pi}{2}\epsilon$ error from the original IAP-rank scores.
Note that computing $\psi$ per datapoint takes time $mm^{\prime}$ since it requires multiplying by the Gaussian matrix $\mathbf{G}$. Also, when $\psi$-transformations are computed, all remaining computations can be conducted in the $m^{\prime}$-dimensional space. That completes the first part of the theorem regarding trigonometric features if we choose $m$ in such a way that $2\exp(-\frac{m\epsilon^{2}}{2})LI \leq p$.

Now we switch to positive random features and softmax sampling.
Without loss of generality we will assume that $L=2^{h}$ for some $h \in \mathbb{N}$.
Note that softmax sampling can be approximated with the following procedure. 
We construct a binary tree $\mathcal{T}$ recursively from its root to its leaves. With every node $v$ we associate a set of $\phi$-transformed keys $\mathcal{S}_{v}=\{\phi(\mathbf{k}^{v}_{1}),...,\phi(\mathbf{k}^{v}_{r_{v}})\} \subseteq \{\phi(\mathbf{k}_{1}),...,\phi(\mathbf{k}_{L})\}$. Whenever $r>1$, we split $\mathcal{S}_{v}$ into two disjoint equal-size subsets: $\mathcal{S}^{\mathrm{left}}_{v}$, $\mathcal{S}^{\mathrm{right}}_{v}$ (in a completely arbitrary way) and create two children of $v$: node $v_{\mathrm{left}}$ with assigned set $\mathcal{S}^{\mathrm{left}}_{v}$ and
node $v_{\mathrm{right}}$ with assigned set $\mathcal{S}^{\mathrm{right}}_{v}$. We associate the set of all $\phi$-transformed keys with the root node of $\mathcal{T}$.
For each node $v$, we also compute $\mathbf{f}(v)=\sum_{\mathbf{k} \in \mathcal{S}_{v}} \phi(\mathbf{k})$. Note that computing $\mathcal{T}$ with all the associated meta-data can be clearly done in time $O(Lm)$ given computed earlier: $\{\phi(\mathbf{k}_{1}),...,\phi(\mathbf{k}_{L})\}$.
Given vector $\mathbf{z}^{\prime}=\phi(\sum_{i=1}^{L}\mathbf{q}_{i})$, we conduct approximate softmax sampling as follows.
We start traversing tree $\mathcal{T}$ in its root. Assume that in time $t$ we are in node $v$ which is not a leaf. We choose as a next node one of its children. We choose the left child with probability:
\begin{equation}
\label{prob}
p_{\mathrm{left}}=\frac{f(v_{\mathrm{left}})^{\top}\mathbf{z}^{\prime}}{f(v_{\mathrm{left}})^{\top}\mathbf{z}^{\prime}+f(v_{\mathrm{right}})^{\top}\mathbf{z}^{\prime}}.
\end{equation}
When we reach the leaf, we output as a result of our approximate softmax sampling procedure a patch of index $i$ corresponding to the set $\{\phi(\mathbf{k}_{i})\}$ in that leaf. Note that this procedure is always well-defined (the probabilities are nonnegative) if we use as $\phi$ a positive feature map from Equation \ref{positive}. Trigonometric features do not guarantee this. Note also that as $m \rightarrow \infty$, the bias of that approximate softmax sampling tends to $0$.
This comes from the fact that the numerator and denominator of the expression on $p_{l}$ from Equation \ref{prob} are unbiased Monte Carlo estimations based on $m$ independent random projections for the positive random feature map from Equation \ref{positive} of the following two expressions:
$\sum_{\mathbf{k} \in \mathcal{R}_{v_{\mathrm{left}}}} \exp((\sum_{i=1}^{L}\mathbf{q}_{i})^{\top}\mathbf{k})$ and
$\sum_{\mathbf{k} \in \mathcal{R}_{v}} \exp((\sum_{i=1}^{L}\mathbf{q}_{i})^{\top}\mathbf{k})$, where
$\mathcal{R}_{v}$ is the set of keys corresponding to vectors from $\mathcal{S}_{v}$. That completes the proof.

\subsubsection{Proof of Theorem \ref{second_theorem}}
\begin{proof}
As in the proof of the previous theorem, we start with the trigonometric random feature part. 
Note that the estimator $\widehat{\mathrm{SM}}_{\mathrm{trig}}(\mathbf{x},\mathbf{y})$ of the softmax kernel based on trigonometric random features from Equation \ref{trig}, can be rewritten as:
\begin{equation}
\widehat{\mathrm{SM}}_{\mathrm{trig}}(\mathbf{x},\mathbf{y})    = \widehat{K}_{\mathrm{gauss}}(\mathbf{x},\mathbf{y})\exp(\frac{\|\mathbf{x}\|^{2}+\|\mathbf{y}\|^{2}}{2}),
\end{equation}
where $\widehat{K}_{\mathrm{gauss}}(\mathbf{x},\mathbf{y})$ stands for the estimator of the Gaussian kernel applying trigonometric features from \citep{rahimi}. Thus, by applying Claim 1 from \citep{rahimi} regarding uniform convergence of trigonometric features, we obtain: 

\begin{equation}
\mathbb{P}\left[\sup_{i,j,t}|\widehat{\mathbf{A}_{t}}(i,j) - \mathbf{A}_{t}(i,j)| \geq \epsilon \right] \leq
2^{8}(\frac{\sqrt{d_{QK}}\mathrm{diam}}{\epsilon})^{2}\exp\left(-\frac{m\epsilon^{2}}{4(d_{QK}+2)}\right)\tau,
\end{equation}
where: $\tau=\exp(\frac{R^{2}}{\sqrt{d_{QK}}})$ and $\mathrm{diam}=2Rd_{QK}^{-\frac{1}{4}}$. In the formula above $\mathbf{A}_{t}$ stands for the attention matrix in step $t$ and $\widehat{\mathbf{A}}_{t}$ is its approximate version obtained implicitly by IAP-rank in step $t$.
Thus the probability that the approximate ranking $\mathbf{r}^{\prime}$ is $\epsilon$-approximate is at least $1 - p$, where 
$p=2^{8}(\frac{\sqrt{d_{QK}}\mathrm{diam}}{\epsilon})^{2}\exp\left(-\frac{m\epsilon^{2}}{4(d_{QK}+2)}\right)\tau$.
Solving this equation for $m$, we obtain the first part of the theorem.
Now let us focus on positive features and queries/keys normalization setting.
Denote by $\widehat{\mathrm{SM}}_{+}(\mathbf{x},\mathbf{y})$ an estimator of the softmax kernel leveraging positive random features. If $\|\mathbf{x}\|_{2}=\|\mathbf{y}\|_{2}=d_{QK}^{\frac{1}{4}}$ then the mean squared error of the estimator, by \citep{performers}, satisfies:
\begin{equation}
\mathrm{MSE}(\widehat{\mathrm{SM}}_{+}(\mathbf{x},\mathbf{y})) = \frac{1}{m}\exp(4\sqrt{d_{QK}}\cos^2(\frac{\theta_{\mathbf{x},\mathbf{y}}}{2}))(1-\exp(-4\sqrt{d_{QK}}\cos^{2}(\frac{\theta_{\mathbf{x},\mathbf{y}}}{2})))\mathrm{SM}^{2}(\mathbf{x},\mathbf{y})  
\end{equation}
Denote by $\mathcal{E}$ an event that in the approximate attention matrix $\widehat{\mathbf{A}}_{\mathrm{K}}$ given by IAP-rank (not explicitly materialized) the entries of the rows corresponding to $\mathcal{P}$ differ from the corresponding entries in the groundtruth attention matrix $\mathbf{A}_{\mathrm{K}}$ by at most $\epsilon$.
By the Chebyshev's inequality and the union bound we have:
\begin{equation}
\mathbb{P}[\mathcal{E}] \geq 1 - L^{2}I\frac{V}{\epsilon^{2}}, 
\end{equation}
where $V = \frac{1}{m}\exp(4\sqrt{d_{QK}}\cos^2(\frac{\pi-\alpha}{2}))(1-\exp(-4\sqrt{d_{QK}}\cos^{2}(\frac{\pi-\alpha}{2})))\mathrm{SM}^{2}(\mathbf{x},\mathbf{y})$.
Note that if $\mathcal{E}$ holds then the scores assigned by IAP-rank to the tokens from $\mathcal{P}$ do not change by more than $\epsilon$ as compared to groundtruth scores. Thus solving $p=L^{2}I\frac{V}{\epsilon^{2}}$ for $m$ we obtain the inequality in the statement of the theorem and complete its proof.
\end{proof}